\pdfoutput=1

\documentclass[11pt]{article}

\usepackage{acl}

\usepackage{times}
\usepackage{latexsym}

\usepackage[T1]{fontenc}

\usepackage[utf8]{inputenc}

\usepackage{microtype}

\usepackage{inconsolata}

\usepackage{url}
\usepackage{graphicx} \usepackage{placeins}
\usepackage{multirow}
\usepackage{hhline}
\usepackage{booktabs}
\usepackage{fancyhdr,amsmath,amssymb}
\usepackage{float}
\usepackage{tabularx}
\usepackage[T1]{fontenc}
\usepackage{tikz}
\usepackage{enumitem}
\usepackage{varioref}

\usepackage[raggedrightboxes]{ragged2e}
\restylefloat{table}
\usepackage[ruled,vlined]{algorithm2e}
\SetInd{0.2em}{0.3em}
\SetArgSty{textup} \usepackage{hyperref} \newcommand{\red}[1]{\textcolor{red}{#1}}

\newcommand{\blue}[1]{\textcolor{blue}{#1}}

\usepackage{marvosym}

\usepackage[usestackEOL]{stackengine}

\title{PRODIGy: a PROfile-based DIalogue Generation dataset}

\author{
 Daniela Occhipinti$^{1,2}$,
 Serra Sinem Tekiro\u{g}lu$^1$,
 Marco Guerini$^1$
 \\
 \ \\
 $^1$Fondazione Bruno Kessler, Via Sommarive 18, Povo, Trento, Italy
 \\
 \texttt{docchipinti@fbk.eu, tekiroglu@fbk.eu, guerini@fbk.eu}
 \\
 $^2$University of Trento, Italy
 \\
}

\begin{document}
\maketitle
\begin{abstract}
Providing dialogue agents with a profile representation can improve their consistency and coherence, leading to better conversations. However, current profile-based dialogue datasets for training such agents contain either explicit profile representations that are simple and dialogue-specific, or implicit representations that are difficult to collect. %
In this work, we introduce the PRODIGy (PROfile-based DIalogue Generation) dataset, which brings %
diverse representations together, providing a more comprehensive profile dimension set for each speaker. This resource comprises more than 20k dialogues, sourced from movie scripts, aligned with speaker representations such as communication style, biography, personality and gender. %
Initial experiments with diverse baselines show that providing generative language models with these aspects of a profile, both separately and jointly, enhances models' performance. This improvement holds true in both in-domain and cross-domain settings, for both fine-tuned and instruction-based LLMs.

\end{abstract}

\section{Introduction}
Dialogue agents capable of holding human-like interactions 
have drawn increasing interest in the fields of AI and NLP, becoming a key topic and challenge in both industry and academia. 
Unlike task-oriented systems focusing on solving specific tasks, open-domain dialogue systems aim to discuss various topics, possibly maintaining a consistent profile in their responses \cite{kann-etal-2022-open}. In this work, we %
investigate the role of profile information in open-domain dialogue systems.

Despite recent advancements in conversational agents, due to the continuous development of neural models \cite{radford2019language, devlin-etal-2019-bert, scao2022bloom, zhang2022opt, peng2022godel}, these agents often struggle to maintain coherence, resulting in inconsistent or uninformative responses. This issue adversely affects user engagement and trust \cite{li-etal-2016-deep, li-etal-2020-dont}. In this scenario, endowing dialogue systems with profile information is crucial for enhancing the models' ability to generate fluent, consistent, and informative responses \cite{li-etal-2016-persona, zhang-etal-2018-personalizing, zemlyanskiy-sha-2018-aiming, song2019exploiting, majumder-etal-2021-unsupervised, mazare-etal-2018-training}.

The concept of \textit{profile} in a dialogue can refer to three aspects: \textit{personalisation}, \textit{persona}, and \textit{personality}. \textit{Personalisation} refers to employing users' information to drive engagement and help them satisfy their needs \cite{vesanen2007personalization}. \textit{Personality}, on the other hand, is a psychological concept meant to capture how we behave and react to the world \cite{allport1937personality, vinciarelli2014survey}. The notion of \textit{persona} can have diverse meanings in literature. In this work, we will stick to the definition provided by \citet{li-etal-2016-persona}, according to which the persona is the character that an artificial agent plays during conversations and includes elements such as background facts, language, and interaction style.

Several approaches have been explored to integrate persona information into dialogue generation \cite{li-etal-2016-persona,mazare-etal-2018-training, welch-etal-2022-leveraging, zhang-etal-2018-personalizing, song-etal-2021-bob, zheng2020pre, cao-etal-2022-model, majumder-etal-2020-like, liu-etal-2020-impress, majumder-etal-2021-unsupervised, zheng2019personalized}. However, these methods are typically sporadic and disjointed, addressing only one persona dimension at a time, either through an \textit{explicit} representation (a few simple, dialogue-specific sentences about the user) or an \textit{implicit} representation (a collection of the user's previous dialogues) that is challenging to obtain. Consequently, these approaches %
fail to model the complex nature of human communication, which is influenced by the interaction of multiple aspects.

\begin{figure}[t]{\includegraphics[width=1\columnwidth]{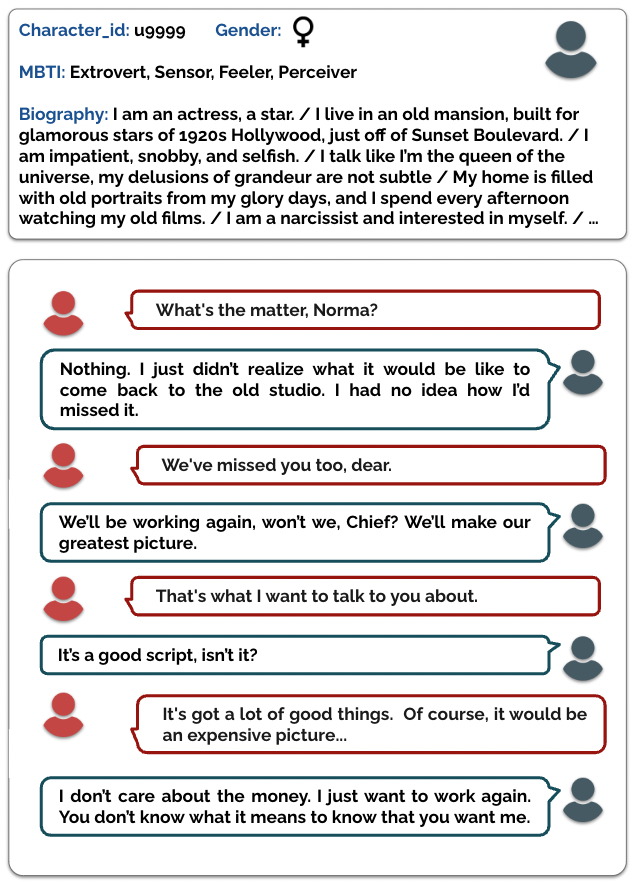}}
 \caption{Example of a dialogue with diverse speaker's profile information provided.}
 \label{fig:persona_pic}
 \end{figure}

In this paper, we investigate the impact of diverse profile representations in the development of dialogue systems by comparing and benchmarking them. %
To this end, we introduce a new dataset, named PRODIGy (PROfile-based DIalogue Generation)\footnote{The dataset will be distributed for research purposes at the following link: \url{https://github.com/LanD-FBK/prodigy-dataset}.
}, that combines existing profile representations (i.e., language style, gender, personality) with novel and more complex representations of the persona, such as biographies. PRODIGy is created starting from the Cornell Movie Dialogs Corpus \cite{danescu-niculescu-mizil-lee-2011-chameleons}, which includes movie script dialogues, and adopting the character IDs and binary gender labels from the original corpus. This approach avoids privacy concerns related to employing real user data and simplifies the distribution. Moreover, the dataset has been aligned with external resources containing characters' profiles, and it can be further expanded by adding new scripts or scripts in other languages. Figure \ref{fig:persona_pic} illustrates %
an example from PRODIGy, in which the dialogue is aligned with the target speaker's profile representation.

We validated PRODIGy by benchmarking it with diverse baselines. In particular, we employed either fine-tuning or instruction prompting, and tested a range of configurations varying the profile dimensions, both in-domain and cross-domain. Evaluation involved both automatic metrics and human assessment.
As for automatic metrics, in-domain experiments show that fine-tuning LMs with diverse profile aspects significantly improves their predictive capabilities. Additionally, instructing non-fine-tuned LLMs with profile information also improves their performance. In cross-domain settings, PRODIGy-based models show better generalisation than those trained on other persona-based resources. 
In human evaluations, evaluators had a tendency of favouring generic responses for broader applicability. However, when responses were consistent with both profile and dialogue they were clearly preferred. Profile information proves beneficial especially in dialogues with limited context, and when disclosed to evaluators, profile-based responses are deemed more appropriate.

\section{Related Work} \label{related-work}

We discuss three main topics relevant to our work: (i) theories on persona and personality (ii) available datasets for persona-based generation and (iii) persona and personality based models.
 
\paragraph{Persona and Personality} \label{persona-personality-theories}

Our communication style is closely related to social status, gender, and motivations, and offers insights into our psychological state \cite{pennebaker2003psychological}. These aspects are closely related to the concepts of \textit{persona} and \textit{personality}, which fall under the more general concept of \textit{profile} \cite{schiaffino2009intelligent}. 
\textit{Persona} can be defined as the character that an artificial agent acts during a conversation and it is a combination of identity factors, such as background facts, language use, and communication style \cite{li-etal-2016-persona}. \textit{Personality} is a psychological concept grasping different behaviours, feelings and way of thinking \cite{allport1937personality, vinciarelli2014survey}. It can be formalised using theoretical frameworks called \textit{trait models}, such as \textit{Big Five} \cite{john1991big} and the \textit{Myers-Briggs Type Indicator (MBTI)} \cite{myers1962myers}. 

\paragraph{Persona-Based Dialogue Datasets} 
Several dialogical datasets contain a persona representation, many of which were collected starting from social media such as Twitter, Reddit, Weibo or Kialo. However, these datasets have various limitations. They may encounter challenges related to ephemerality \cite{klubicka2018examining}; they can include short conversations, thus failing to fully represent real dialogues \cite{li-etal-2016-persona, mazare-etal-2018-training}; they can rely only %
on users' dialogue history \cite{qian2021pchatbot}; they may include only generic persona representations such as gender or age \cite{zheng2019personalized, zhong-etal-2020-towards}; finally, they may not consider linguistic style, being based on controlled and redacted conversations \cite{scialom-etal-2020-toward}. Other resources were collected from television series transcripts \cite{li-etal-2016-persona}, but are small and %
not sufficient to train open-domain dialogue models. One of the most widely used persona-based datasets is Persona-Chat \cite{zhang-etal-2018-personalizing}, collected in a controlled crowd-sourcing environment. However, it provides %
a generic fact-based persona representation (e.g. \textit{"I just got my nails done"}) %
specific to single dialogues and leaving out complex aspects, such as linguistic style or biographical history. 

\paragraph{Persona/Personality Based Dialogue Models} Several approaches have been investigated to condition the dialogue generation through the persona information. On the one hand, diverse studies were based on resources %
using users' past dialogues to represent the persona \cite{li-etal-2016-persona, mazare-etal-2018-training, zhong-etal-2020-towards}.
On the other hand, a line of research has been built on Persona-Chat. Various approaches employed this dataset to train persona-based models in under-resourced scenarios \cite{song-etal-2021-bob, zheng2020pre, cao-etal-2022-model}.
Other methodologies used Persona-Chat to test commonsense expansion \cite{majumder-etal-2020-like}, mutual perception persona \cite{liu-etal-2020-impress}, or enriching persona information through background stories \cite{majumder-etal-2021-unsupervised}. However, these studies present the same limitations of the resources they rely on.
Regarding the personality-driven generation, few seminal studies have been conducted \cite{mairesse-walker-2007-personage, mairesse2008personality, gill-etal-2012-perceptions}.  However, they leave the interactions between personality and persona unexplored.

\section{Construction of the PRODIGy dataset} \label{proposed-solution}

To build the PRODIGy dataset, we started from the Cornell Movie Dialogs Corpus, a dataset of dialogues from movie scripts that includes metadata about movie genre, release year and characters' gender \cite{danescu-niculescu-mizil-lee-2011-chameleons}. The dialogues in the Cornell Movie Dialogs Corpus are between two actors and have an average length of 4 turns. 
The reason for using this resource as a starting point is three-fold: (i) \textit{Data Persistency and Accessibility:} it eliminates privacy issues or ephemerality problems \cite{klubicka2018examining} that would arise from collecting data from real users and, therefore, facilitates the distribution of PRODIGy to the research community; (ii) \textit{Data Enrichment:} it is possible to enrich PRODIGy with the profile of movie characters through the alignment with external web resources containing information about characters and movie plots; 
(iii) \textit{Data Expansion:} it leaves room for further development/extension; for example, it can be aligned with similar movie script resources in other languages or new movie scripts. 

Below, we outline the profile representations and detail the methodology employed to annotate the characters within the dataset.

\paragraph{Dialogical Information.} Following previous approaches \cite{li-etal-2016-persona, qian2021pchatbot}, we provide an implicit representation of each character's persona through a collection of characters' dialogues. Thus, we can represent the characters' linguistic styles. To this end, we included in PRODIGy only the characters with at least 50 dialogues in the Cornell Movie Dialogs Corpus.
 
\paragraph{Personality Information.} To associate each character with \textit{personality} information, we cross-referenced the Cornell Movie Dialogs Corpus with the Personality Database (PDB)\footnote{\href{https://www.personality-database.com/}{https://www.personality-database.com/}} website. PDB is a widely used social platform in which users can assign personality types from several trait models to fictional characters and real famous people. We use this platform as a provider of crowd-sourced characters' personality annotations.

To annotate the characters in the Cornell Movie Dialogs Corpus, we 
used the query \texttt{movie\_title+year} to extract from PDB the metadata related to each movie, containing the list of the characters' names and IDs. If the character was present in the metadata, we used the query \texttt{PDB\_characterID} to extract the MBTI type and related votes. If the MBTI type had at least 5 votes, the character was annotated. If the character was not in the metadata, a human annotator performed a manual check within PDB to verify if there was an actual match. In case the mismatch could be manually resolved, we replicated the above procedure to annotate the character. 
Details of the alignment procedure are provided in Appendix \ref{annotation-mbti-appendix}.

Among the several trait models provided by PDB on each character's web page, we focused on MBTI since it is widely studied and it was the most voted model by users, thus proving a more stable and reliable crowd-annotation. The MBTI trait model takes into account 16 personality types obtained from the combination of 4 dichotomies: introversion or extroversion, sensing or intuition, thinking or feeling, and judging or perceiving.

In line with the definition of personality traits, which posits their stability over time, we assigned a unique MBTI personality type to each character. This differs from the approach of \citet{jiang2020automatic}, who assigned a different personality for each dialogue in which the character is present. Finally, %
for annotation reliability, we discarded the characters (and related dialogues) with less than 5 user votes and used the personality type derived from the majority of votes on each MBTI dichotomy.
 
\paragraph{Biographical Information.} The third step was to provide the characters with explicit persona representations that serve as background information for all the dialogues in which the character is present. Inspired by the concept of \textit{background story} by \citet{majumder-etal-2021-unsupervised}, we aim to provide a representation that goes beyond simple facts. To this end, we consider the biographical information. %
We scraped the biographies of the characters annotated with the personality information, from Charactour.com, Fandom.com and Wikipedia. Then, to automatically extract the most relevant sentences, we employed %
an extractive summarisation algorithm based on Kullback-Leibler distance \cite{haghighi-vanderwende-2009-exploring}. Subsequently, a human-machine collaboration procedure followed, where a human annotator\footnote{The human annotator was one of the authors and a Computer Science PhD student.} 
modified the extracted sentences to ensure that our resource maintains %
an alignment with the Persona-Chat dataset \cite{zhang-etal-2018-personalizing} for comparability purposes. To achieve this, specific guidelines were formulated and provided to the annotator:

\begin{itemize}[leftmargin=*]
\itemsep-0.3em
\item Re-rank the top 10 sentences in order of importance, according to the speaker's profile.
\item Convert the sentences from the third to the first person singular.
\item Shorten excessively long sentences.
\item Enrich the sentences with missing relevant information;
\item If a character biography was not found, create one by reading the movie plot.
\end{itemize}

In particular, the annotator re-ranked the sentences giving priority to crucial information that the summarisation algorithm might have originally positioned towards the end of the list, ensuring it now appears within the top five sentences. The importance criterion %
followed the structure of a small selection of biographies that were considered as gold. For instance, details about characters' job, lifestyle, or family background were expected to be on top of the list. %

While PRODIGy biography sentences align stylistically with Persona-Chat \cite{zhang-etal-2018-personalizing}, they are not limited to generic facts and capture more complex aspects of the persona, making them qualitatively different from Persona-Chat.

To increase the number and the variability of biography sentences, ChatGPT \cite{openai-chatgpt} was given the original sentences and asked to produce two paraphrases. These new sentences were given to the annotator for post-editing to correct errors or further paraphrase those still too similar to the original biographies. %
More details about the biographical information procedure are provided in Appendix \ref{annotation-biography-appendix}. In Table \ref{tab:bio_example}, we present an example of the biography editing process.

\begin{table*}[ht!]
    \small
    \centering
    \begin{tabular}{p{0.30\linewidth}|p{0.30\linewidth}|p{0.30\linewidth}}
        \toprule
        \multicolumn{1}{c|}{\textbf{Extracted bio}} & \multicolumn{1}{c|}{\textbf{Post-edited bio}} & \multicolumn{1}{c}{\textbf{Paraphrased bio}}\\
        \midrule
        \vspace{-0.55cm}
        \begin{enumerate}[leftmargin=*]
        \itemsep-0.3em
            \item \blue{He is too young to be so sick.}
            \item Living... in Vienna with his beautiful wife Constanze and their young son.
            \item Relationship Status... on the rocks. He loves Constanze, but he is not making her happy.
            \item They say he can't be trusted with young girls.
            \item \red{Profession... composer.}
        \end{enumerate}
        &
        \vspace{-0.55cm}
        \begin{enumerate} [leftmargin=*]
        \itemsep0em
            \item \red{I am a composer.}
            \item I live in Vienna with my beautiful wife Constanze and our young son.
            \item My relationship is on the rocks: I love my wife Constanze, but I am not making her happy.
            \item \blue{I am too young to be so sick.}
            \item They say I can't be trusted with young girls.
        \end{enumerate} 
        &
        \vspace{-0.55cm}
        \begin{enumerate} [leftmargin=*]
        \itemsep0em
            \item I am a musician who specializes in composition.
            \item I live in Vienna with my wife Constanze and our young son.
            \item My relationship with Constanze is strained: I love her, but I am not making her happy.
            \item I am too young to be suffering from illness.
            \item People say that I cannot be trusted around young girls.
        \end{enumerate}\\
        \bottomrule
    \end{tabular}
    \caption{Example of the modifications made to a biography during the editing process, along with one of the corresponding paraphrases. Colour highlights indicate sentences that were re-ranked (e.g., a sentence ranked 6th in Extracted Bio is moved to 1st position in Post-Edited Bio).}

    \label{tab:bio_example}
\end{table*}

As a result of the aforementioned procedures, we obtained a dataset with more than 20K dialogues for 80K turns with 300 annotated characters and more than 8k biography sentences. The dialogues are aligned with the following dimensions of one of the speakers: gender, personality type, character's biography, and linguistic style modelled by character's dialogues. %
Character biographies consist of an average of 8 sentences, ranging from 5 to 10 sentences, with an average of 13 tokens per sentence. %
Each biography sentence has been paraphrased twice.
Detailed statistics of the PRODIGy dataset are provided in Table \ref{dataset-stats}. 

\begin{table}[]
\centering
\small
\begin{tabular}{l|r} 
\toprule
\textbf{Category} & \textbf{Statistics} \\
\midrule
Dialogues & 20850 \\
Turns & 80604 \\
Annotated Characters & 339 \\ 
Biography Sentences & 8498 \\ 
\midrule
Turns per Dialogue & 4 ($\pm$3.28)\\
Dialogues per Character & 78 ($\pm$31.21)\\
Sentences per Bio & 8 ($\pm$1.57) \\ 
Token per Bio Sentence & 13 ($\pm$5.66)\\
\bottomrule
\end{tabular}
\caption{PRODIGy main statistics. The upper part reports counts, while the lower reports averages.}
\label{dataset-stats}
\end{table}

\section{Baselines and Experiments} \label{persona-personality-generation}

In this section, we propose several configurations to condition the dialogue generation with profile information. In particular, we represent profiles by using either the persona, the personality information, or both. Our aim is to analyse the impact of each representation on the generation process. 

For all the configurations, we employed the DialoGPT model as our baseline since it is a generative transformer-based model pre-trained on conversation-like exchanges \cite{zhang-etal-2020-dialogpt}, making it the most suitable baseline for the dialogue generation task. %
We investigated several fine-tuning configurations. As a baseline, we fine-tuned DialoGPT without any profile information, while in the remaining configurations we fine-tuned the model considering both single profile dimensions and their combinations. Specifically, we concatenated the characters' profile information to the corresponding turns of the dialogues. In Appendix \ref{dialogpt-training-setup-appendix}, we provide details on the fine-tuning setup and input syntax utilised for DialoGPT.

Besides DialoGPT, we also experimented with GODEL \cite{peng2022godel}, an instruction-based LLM specific for dialogue generation. Our aim is to assess the effect of providing %
profile information %
as an instruction to a non-fine-tuned LLM. The input syntax for GODEL is shown in Appendix \ref{GODEL-input-syntax-appendix}.

Although more powerful models are available, such as ChatGPT \cite{openai-chatgpt} and LLaMa 2-chat \cite{touvron2023llama}, we chose to use DialoGPT and GODEL as our baselines for the following reasons: (i) ChatGPT and LLaMa 2-chat are explicitly intended for assistant-like chat (i.e. human-machine interactions), whereas our goal is to explore dialogue models simulating broader human-human interactions, playing the role of any of the two speakers; (ii) we chose two language models comparable in pre-training data (i.e., similar human-human dialogical interactions) and parameter size; (iii) these models were already used for the dialogue generation task and allow testing of two main approaches: DialoGPT for fine-tuning and GODEL for instruction prompting in a zero-shot setting.

Regarding the inspected configurations, we provide the description as follows: 
\paragraph{Plain Dialogue Driven Generation} \label{plain-dialogue-gen}
In the first configuration, we fine-tuned DialoGPT and instructed GODEL only with the plain dialogue, without considering any profile information. This configuration will be used as a baseline to assess the improvement obtained by adding the various profile information to both models. 
\paragraph{Personality Driven Generation} \label{personality-generation}
In this configuration, we employ PRODIGy and the characters' MBTI to fine-tune DialoGPT and prompt GODEL, as it is possible to generate language reflecting a certain personality type %
\citep{mairesse-walker-2007-personage, mairesse2008personality,gill-etal-2012-perceptions}. 
\paragraph{Persona Driven Generation} \label{persona-generation}
In this configuration, we employ the implicit (i.e. linguistic and stylistic information) and explicit (i.e. gender and biography sentences) persona representations in PRODIGy, either individually or jointly. This enabled us to analyse the effect of each representation and combination in the dialogue generation.\\

Firstly, we used the characters' dialogues as implicit persona representation \cite{li-etal-2016-persona, qian2021pchatbot}. We fine-tuned DialoGPT on PRODIGy, %
aggregating characters' dialogue lists using their IDs to capture their linguistic styles. Secondly, inspired by \citet{zheng2019personalized} and \citet{schwartz2013personality}, we considered gender as another persona representation to fine-tune DialoGPT and instruct GODEL. Then, motivated by \citet{zhang-etal-2018-personalizing}, %
we provided DialoGPT and GODEL with persona information in the form of biography sentences. Our aim is to generate non-generic and informative responses that are consistent with both the dialogues and the biography sentences. 

\paragraph{Inter-Character and Intra-Character Configurations}

Using PRODIGy, we set up two configurations to train DialoGPT: \textit{inter-character} and \textit{intra-character}. In the first configuration, the test characters are not used at training time. In the second configuration, at training time the system learns about the specific characters to be predicted at test time. In both cases, we use only 5 biography sentences, following %
\citet{zhang-etal-2018-personalizing}. These two configurations %
also address privacy concerns: %
in one case, the LM does not retain any personal information but uses it only at inference time, while in the second, the LM stores the information about the user in its internal representation.

\section{Automatic Evaluation}

In this section, we describe the metrics %
and experiments for the validation of our resource.

\subsection{Metrics} \label{metrics}

We assess model performances using two metrics: \textit{Conditional turn Perplexity} \cite{su-etal-2021-put} and \textit{Average Accuracy at N} \cite{welch-etal-2022-leveraging}. 

Conditional Perplexity (\begin{math}CPPL\end{math}) in our scenario is the perplexity of a gold turn given the context. \begin{math}CPPL\end{math} %
is used to compute the model likelihood of a turn given a dialogue history and possible profile information (see Appendix \ref{cppl-appendix} for the formulation).  %
With Average Accuracy at \begin{math}N\end{math} (Acc@\begin{math}N\end{math}), the prediction of a word from a gold turn is considered correct if it occurs within the top \begin{math}N\end{math} most probable words given by the model. %

We adopted these metrics to evaluate our models in both in-domain (i.e., on PRODIGy) and cross-domain (i.e., on Persona-Chat) scenarios.

\subsection{Analysis and Results} \label{initial-work}
 
In this section, we provide a detailed description of the following experiments: (i) Inter-Character Experiments, (ii) Intra-Character Experiments, (iii) Cross-Domain Experiments. In these %
settings, we consider the target speaker's profile, excluding the interlocutor' profile. Given just the dialogue context, or both context and profile information, we aim to predict the target speaker's final turn. 

\paragraph{Inter-Character Experiments}

In this setting, we partitioned PRODIGy making sure that the characters in the test set are not present in the training set, consistently with the experiments by \citet{welch-etal-2022-leveraging}. We opted for the {Bio}$_{par}$ model as our biography-based model. This model is trained by randomly selecting five sentences\footnote{We employ only 5 biography sentences to ensure (i) we stay within the DialoGPT input size length of 1024 tokens, (ii) we are consistent with Persona-Chat configuration.} per dialogue from the original biography or its paraphrases. The decision to use this model is based on its demonstrated superior effectiveness, as shown in a preliminary experiment (outlined in Appendix \ref{bio-experiment}) focusing on biography-based models.

Table \ref{tab:PRODIGy-results} presents model performances based on profile information. In terms of Acc@\begin{math}N\end{math}, these models outperform Plain Dialogue that lacks profile information. Single-profile models show similar Acc@10 performances. Also, combining multiple profile dimensions, the Acc@\begin{math}N\end{math} scores do not differ significantly. Regarding \begin{math}CPPL\end{math}, Plain Dialogue performs the worst, while models with profile information excel. Notably, Gender attains the best \begin{math}CPPL\end{math} (87.92), comparable to MBTI. Bio$_{par}$ performs worse than Gender and MBTI but significantly outperforms the baseline with a score of 98.27, showcasing the efficacy of high-level character descriptions. Gender's strong performance in \begin{math}CPPL\end{math} and Acc@\begin{math}N\end{math} may stem from the gender-specific linguistic patterns in PRODIGy's dialogues sourced from the Cornell Movie Dialogs Corpus \cite{schofield-mehr-2016-gender}, enabling the model to effectively incorporate such characteristics. Overall, the results show that adding profile information, either alone or jointly, strongly improves the models performance in terms of generalisation\footnote{Besides \begin{math}CPPL\end{math} and Acc@\begin{math}N\end{math}, we explored coherence and groundedness metrics. Results, detailed in Appendix \ref{coherence-analysis}, align with the main findings with profile-based models performing better than plain dialogue model.}.

\begin{table}[ht!]
\small
\centering
\begin{tabular}{@{}lrrrrrr@{}}
\toprule
 \textbf{Config.} & \textbf{CPPL} %
 & \textbf{Acc@10} & \textbf{Acc@1} \\ \midrule
\texttt{MBTI} & 89.30 %
& \textbf{0.665} & \textbf{0.317} \\
\Hermaphrodite & \textbf{87.92} %
& 0.664 & 0.306 \\
Bio$_{par}$ & 98.27 %
& 0.661 & 0.307 \\

PD & 541.16 %
& 0.585 & 0.298 \\
 [+1ex]
\hline 
\\[-1ex]
\texttt{MBTI}+\Hermaphrodite & \textbf{91.50} %
& \textbf{0.660} & \textbf{0.311} \\
\Hermaphrodite+Bio$_{par}$ & 96.31 %
& 0.658 & 0.299 \\
\texttt{MBTI}+Bio$_{par}$ & 100.35 %
& 0.653 & 0.296 \\
 [+1ex]
\hline 
\\[-1ex]

\texttt{MBTI}+\Hermaphrodite+Bio$_{par}$ & 91.65 %
& 0.660 & 0.302
 \\ \bottomrule
\end{tabular}
\caption{DialoGPT results on PRODIGy test set (Inter-Character). PD and \Hermaphrodite represent Plain Dialogue and Gender, respectively.}
\label{tab:PRODIGy-results}
\end{table}

In Table \ref{tab:godel-results} we report the results obtained by prompting GODEL with the profile information. %
The \begin{math}CPPL\end{math} and Acc@\begin{math}N\end{math} values reveal better performances even when profile information is merely provided as an instruction. In particular, Plain Dialogue exhibits a worst \begin{math}CPPL\end{math} compared to MBTI and MBTI + Gender (24.00 vs 12.46).
Also in terms of %
Acc@10, MBTI + Gender turned out to be the best-performing model. In terms of Acc@1, the best performing models are Bio and Plain Dialogue, with a score of 0.027, although they do not yield much better performances than the other models. These results show that profile information is beneficial also when prompted to non-fine-tuned instruction-based LLMs. It is important to state that, while GODEL may seem to outperform DialoGPT in terms of \begin{math}CPPL\end{math}, a direct comparison between their metrics is not possible as these models are pre-trained on distinct datasets and have a different vocabulary size.

\begin{table}[ht!]
\small
\centering
\begin{tabular}{@{}lrrrr@{}}
\toprule
 \textbf{Config.} & \textbf{CPPL} & \textbf{Acc@10} & \textbf{Acc@1} \\ \midrule
 \texttt{MBTI} & \textbf{12.46} & 0.080 & 0.026 \\
\Hermaphrodite & 13.65 & 0.075 & 0.026 \\
Bio & 20.43 & \textbf{0.082} & \textbf{0.027} \\
PD & 24.00 & 0.074 & \textbf{0.027} \\ 
[+1ex]
\hline 
\\[-1ex]
\texttt{MBTI} + \Hermaphrodite &\textbf{12.46} & \textbf{0.083} & 0.025  \\
\texttt{MBTI} + Bio & 26.48 & \textbf{0.083} & \textbf{0.026}  \\
\Hermaphrodite + Bio & 22.50 & 0.081 & \textbf{0.026}  \\ 
[+1ex]
\hline 
\\[-1ex]
\texttt{MBTI} + \Hermaphrodite + Bio & 28.96 & 0.083 & 0.026  \\ 
[+1ex]
\hline 
\\[-1ex]
\end{tabular}
\caption{GODEL results on PRODIGy test set (Inter-Character). PD and \Hermaphrodite represent Plain Dialogue and Gender, respectively.}
\label{tab:godel-results}
\end{table}

\paragraph{Intra-Character Experiments}
In the second set of experiments, we partitioned PRODIGy with the same character existing in both training and test sets. Our aim is to simulate a scenario in which we can access the information about a character already at training time, both explicitly (i.e. MBTI, gender, and biography) and implicitly (i.e. the character's dialogues, captured by the character ID, grasping their language style).

As shown in Table \ref{tab:intra-characters-results}, endowing the model with the dialogical information (\texttt{ID}) provides the best results in terms of \begin{math}CPPL\end{math}. %
This is attributed to the model learning the character's vocabulary and language style during training, enhancing predictions.
In terms of Acc@\begin{math}N\end{math}, the best performing model is Bio (0.712 of Acc@10, and 0.348 of Acc@1). The other profile-based models exhibit similar performances. The Plain Dialogue model emerges as the weakest, %
proving again that fine-tuning models through profile information is beneficial. %
Combining biographical information and \texttt{ID} further enhances model efficiency in terms of \begin{math}CPPL\end{math}, with better values when a high-level character description is included. 
The scores in Acc@\begin{math}N\end{math} show that, when combined with the dialogical information (\texttt{ID}), the biographical information improves the predictive ability of the model more than Gender and MBTI. Although \texttt{ID} excels in \begin{math}CPPL\end{math}, models with explicit profile information show comparable efficiency. %
Regarding the models trained with profile information jointly, the best performances are achieved by those trained with the characters' biographical information. Generally, models perform better in the Intra-Character setup than in the Inter-Character since they are trained with the speaker's profile information and leverage it at test time.

\begin{table}[ht!]
\small
\centering
\begin{tabular}{@{}lrrrr@{}}
\toprule
 \textbf{Config.} & \textbf{CPPL} & \textbf{Acc@10} & \textbf{Acc@1} \\ \midrule

Bio & 58.95 & \textbf{0.712} & \textbf{0.348} \\
\texttt{ID} & \textbf{55.25} & 0.709 & 0.345 \\
\Hermaphrodite & 58.32 & 0.706 & 0.335 \\
\texttt{MBTI} & 58.32 & 0.706 & 0.346 \\
PD & 595.14 & 0.368 & 0.337 \\
[+1ex]
\hline 
\\[-1ex]
\texttt{ID}+Bio & \textbf{54.89} & \textbf{0.714} & \textbf{0.347} \\
\texttt{ID}+\Hermaphrodite & 58.88 & 0.706 & 0.337 \\
\texttt{ID}+\texttt{MBTI} & 57.82 & 0.704 & 0.343 \\
\Hermaphrodite+Bio & 55.73 & 0.708 & 0.343 \\
\texttt{MBTI}+Bio & 55.95 & 0.708 & 0.344 \\
\texttt{MBTI}+\Hermaphrodite & 58.32 & 0.704 & \textbf{0.347} \\
[+1ex]
\hline 
\\[-1ex]
\texttt{MBTI}+\Hermaphrodite+Bio & 57.08 & \textbf{0.710} & 0.339 \\
\texttt{ID}+\texttt{MBTI}+Bio & \textbf{53.23} & 0.710 & 0.340 \\
\texttt{ID}+\texttt{MBTI}+\Hermaphrodite & 55.48 & 0.705 & \textbf{0.344} \\
[+1ex]
\hline 
\\[-1ex]
\texttt{ID}+\texttt{MBTI}+\Hermaphrodite+Bio & 54.99 & 0.710 & 0.341 \\
\bottomrule
\end{tabular}
\caption{DialoGPT results on PRODIGy test set (Intra-Character). PD and \Hermaphrodite represent Plain Dialogue and Gender, respectively. }
\label{tab:intra-characters-results}
\end{table}

\paragraph{Cross-Domain Experiments}

To evaluate the generalisation capabilities of the models trained on the PRODIGy dataset in a cross-domain scenario, we also analysed the model performances, trained both with no profile information and with biographical information, on the Persona-Chat test set \cite{zhang-etal-2018-personalizing}. These results are also compared with the models trained with the same methodology on Persona-Chat and tested on the PRODIGy test set. 
The results, presented in Table \ref{tab:persona-chat-results}, show a significant improvement in \begin{math}CPPL\end{math} scores when incorporating biography sentences, even in zero-shot settings (both trained on PRODIGy and tested on Persona-Chat, and vice-versa). Interestingly, using a general biography, as the one we propose, yields better generalisation capabilities than a dialogue-specific persona as in \citet{zhang-etal-2018-personalizing}. %
When models trained on PRODIGy are tested on Persona-Chat, the results are in line with the in-domain experiments: Bio$_{par}$ consistently outperforms Plain Dialogue in both \begin{math}CPPL\end{math} and Acc@\begin{math}N\end{math}.
On the contrary, in the scenario in which we trained the models on Persona-Chat and tested on PRODIGy, the Bio model's Acc@\begin{math}N\end{math} scores are lower than Plain Dialogue's scores. This might suggest that persona sentences do not capture personas' complex characteristics, therefore they might be less effective to generalise in a cross-domain scenario.

\begin{table}[ht!]
\small
\centering
\begin{tabular}{@{}llrrr@{}}
\toprule
 \textbf{Train $\rightarrow$ Test} & \textbf{Config.} & \textbf{CPPL} & \textbf{Acc@10} & \textbf{Acc@1} \\ \midrule 
\multirow{2}{*}{PROD. $\rightarrow$ PC} & PD & 891.80 & 0.444 & 0.184 \\
& Bio$_{par}$ & \textbf{219.07} & \textbf{0.533} & \textbf{0.200} \\ [+1ex]
\hline \\[-1ex]
\multirow{2}{*}{PC $\rightarrow$ PROD.}&PD & 1.32\begin{math}e\end{math}+05 & \textbf{0.333} & \textbf{0.139} \\
& Bio & \textbf{3.27\begin{math}e\end{math}+04 }& 0.309 & 0.119 \\
\bottomrule
\end{tabular}
\caption{DialoGPT results on cross-domain experiments: fine-tuning on PRODIGy and test on Persona-Chat (PROD. $\rightarrow$ PC) and vice-versa (PC $\rightarrow$ PROD.). PD represents Plain Dialogue.}
\label{tab:persona-chat-results}
\end{table}

\section{Human Evaluation} \label{human-eval}

Besides the automatic evaluation, we also run an  human evaluation study to validate PRODIGy.

This evaluation involved six subjects, comprising four PhD students in Computer Science and two MSc students in Data Science. %
Evaluators received 100 dialogues each, 50 with profile information disclosed and 50 without profile disclosure, so to enable an assessment of profile information's impact on judgements. %
We focused on output generated using top-p decoding by four models trained during inter-character experiments: the model trained on dialogues only and the models trained with one profile dimension. Evaluators ranked five possible responses for each dialogue, including the gold response used as a control condition, on a scale from 1 (most likely) to 5 (least likely) based on perceived likelihood of being the target speaker's response.
In total, we collected 3000 evaluations. Subsequently, we conducted post-hoc qualitative interviews with the evaluators.

\subsection{Results}\label{human-eval-results}

The human evaluation reveals that the gold responses are preferred by far over the generated responses, indicating clear room for future improvement over the baselines we employed. Notably, Plain Dialogue was the favoured model, with only marginal rating differences compared to other models. From the post-hoc interviews, it emerged that Plain Dialogue's ability to produce generic responses that easily fit into various dialogues was often the reason for this preference. However, an interesting shift occurs when evaluators are made aware of the speaker's profile. In such cases, there is a noticeable increase in the preference for profile-based model responses over Plain Dialogue responses. This shift is shown in Table \ref{tab:profile-vs-plaindialogue}, which outlines the percentages of times evaluators favored profile-based models over Plain Dialogue. 
This trend can be attributed to a clear preference towards generations that exhibit coherence with both profile information and dialogue context, emphasising the significance of the profile in the generation process. Finally, profile-based models receive more favourable evaluations in shorter contexts, suggesting that the inclusion of profile information is advantageous when the dialogue context provides limited information about the speaker.

\begin{table}[ht!]
\small
\centering
\begin{tabularx}{\linewidth}{l *{6}{>{\centering\arraybackslash}X}}
\toprule
 & \multicolumn{2}{r}{\textbf{All turns}} & \multicolumn{2}{r}{\textbf{$\leq $ 6 turns}} & \multicolumn{2}{r}{\textbf{$>$ 6 turns}} \\
\cmidrule(lr){2-3} \cmidrule(lr){4-5} \cmidrule(lr){6-7}
\textbf{Response} & \textbf{No \includegraphics[height=1em]{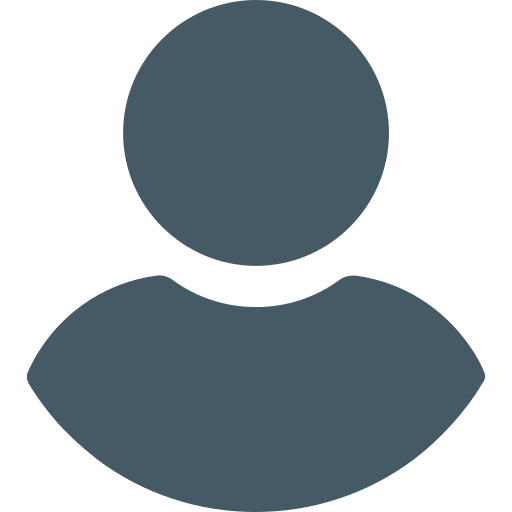}} & \textbf{With \includegraphics[height=1em]{profile_pic.png}} & \textbf{No \includegraphics[height=1em]{profile_pic.png}} & \textbf{With \includegraphics[height=1em]{profile_pic.png}} & \textbf{No \includegraphics[height=1em]{profile_pic.png}} & \textbf{With \includegraphics[height=1em]{profile_pic.png}} \\
\midrule
Bio$_{par}$ & 43.14 & 47.60 & 44.30 & 47.85 & 40.95 & 47.14 \\
MBTI & 44.96 & \textbf{49.59} & 46.33 & \textbf{50.38} & 42.38 & 48.10 \\
\Hermaphrodite & 45.36 & 44.04 & 46.19 & 43.91 & 44.29 & \textbf{49.52} \\
\bottomrule
\end{tabularx}
\caption{Preference Percentages across different dialogue lengths: responses of profile-based Models vs. Plain Dialogue Responses. \Hermaphrodite represents Gender. No/With \includegraphics[height=1em]{profile_pic.png} indicates profile information disclosure to evaluators.}
\label{tab:profile-vs-plaindialogue}
\end{table}

\begin{table*}[ht!]
\centering
\small
\begin{tabular}{p{0.45\textwidth}|p{0.45\textwidth}}
\toprule
\textbf{Input} & \textbf{Responses} \\
\midrule
\begin{minipage}[t]{0.45\textwidth}
\textbf{\texttt{Biography}}: (1) I am a janitor at the Massachusetts Institute of Technology. (2) I live in a rough, working-class Irish neighborhood in south Boston. (3) I am seeing a Harvard girl named Skylar. (4) I am cocky, brash, and combative: while I am often crude, my quick wits mark me as unmistakably intelligent. (5) I have to put my mathematical genius to work by helping Professor Gerald Lambeau and start seeing a psychiatrist. \\
\textbf{\texttt{MBTI}}: Introvert, Intuitive, Thinker, Perceiver \\
\textbf{\texttt{Gender}}: Male \\
\vspace{0.15cm}
\hrule
\vspace{0.15cm}
\textbf{\texttt{Dialogue}:}\\
  \textbf{A:} Okay, you’re in your bed. Now how old are you? \\
  \textbf{B:} Seven. \\
  \textbf{A:} And what do you see? \\
  \textbf{B:} Somethin’s in my room. \\
  \textbf{A:} What is it? \\

\end{minipage}
& 
\begin{minipage}[t]{0.45\textwidth}
\begin{itemize}[leftmargin=*]
\itemsep0em
  \item \textbf{[\texttt{Gold}]} \textbf{B:} It’s like a small figure, hoverin’ over me. Gettin’ closer.
  \item \textbf{[\texttt{Plain Dialogue}]} \textbf{B:} I don’t know. It just looks like a black couch. It’s too bright.
  \item \textbf{[\texttt{Bio$_{par}$]}} \textbf{B:} I see a skeleton.
  \item \textbf{[\texttt{MBTI}]} \textbf{B:} I’m sure you can tell me.
  \item \textbf{[\texttt{Gender}]} \textbf{B:} It’s a door, I don’t know which one but I want to find out for sure and put an end to this shit before it gets any worse.
\end{itemize}
\end{minipage} \\
\bottomrule
\end{tabular}
\caption{Example of Inter-Character models' generations. %
}
\label{tab:example-generations}
\end{table*}

While Plain Dialogue might be favoured for its generic responses adaptable to various dialogues, it is worth noting that each profile-based model learns unique patterns from the profile information during training, resulting in responses tailored to individual speakers. The example in Table \ref{tab:example-generations} illustrates this phenomenon. Plain Dialogue's response is a fairly generic answer that fits the context of the dialogue well. However, each profile-based model's generation reflects the speaker’s profile information. Bio$_{par}$'s output closely aligns with the Gold response concept. Given the character's biography indicating a need for psychiatric help, the model inferred a potential mental distress, responding with "\textit{I see a skeleton.}". The MBTI response aligns with the introverted trait of the character, who is reluctant to answer the interlocutor: "\textit{I'm sure you can tell me.}". The Gender model's response incorporates stereotypical male patterns (e.g. the use of the swear word "\textit{shit}"), common in the Cornell Movie Dialogs corpus \cite{schofield-mehr-2016-gender}. 

These findings are consistent with the feedbacks from evaluators that we gathered in a post-hoc interview. Evaluators expressed a preference for generic answers, typically generated by Plain Dialogue, due to their broader applicability. This was particularly evident in those cases where responses generated by profile-based models matched the profile information of the speaker but not the dialogue context, thus negatively impacting perceived answer quality. However, when profile information was provided to evaluators, the preference for responses consistent %
with both profile and dialogue clearly emerged. At a closer inspection of such cases, we found that these sentences, consistent with both profile and dialogue, were often preferred even to gold responses. %
Conversely, the overarching inclination for gold responses was not given because they were familiar to evaluators: they reported not recognising them, and more broadly to having seen only few of the movies whose dialogues were evaluated. See Appendix \ref{human-evaluation-appendix} for additional details.%

\section{Conclusion} \label{conclusion}

In this paper we introduced PRODIGy, a new dataset of movie dialogues aligned with characters' profile information, i.e. personality type, gender, biography, and a collection of speakers' dialogues, useful for inferring their vocabulary and language style. Derived from movie scripts, PRODIGy also mitigates privacy concerns associated with real user data. To validate this resource, we conducted several experiments using diverse baselines, both via fine-tuning and instruction prompting. %
Results indicate that including profile information in both approaches improved models' performance. Moreover, the cross-domain experiments showed that PRODIGy-based models exhibit better generalisation than those trained on similar resources. Results from the human evaluation showed that, despite a preference for generic responses due to their broader applicability, responses consistent with both profile and dialogue are clearly favoured. Moreover, the results highlight the value of incorporating profile information, especially when speaker's information provided within the dialogue context is limited.

\section*{Limitations} \label{limitations}

The fact that PRODIGy includes fictional characters could imply that the roles may be stereotyped. The high predictivity of the model trained on characters' gender is a potential indicator of this hypothesis. Thus, while PRODIGy allows avoiding a number of privacy issues, it may be less realistic. However, this problem may be present in other datasets, such as Persona-Chat, where users were simulated. Moreover, as regards to Gender, PRODIGy is limited to a binary classification since it is the one originally provided by the Cornell Movie Dialogs Corpus. Finally, the human evaluation shows a strong preference for gold responses, suggesting significant room for improvement, which we plan to address in future work.

\section*{Ethics Statement} \label{ethics}

One of the potential risks of profile-based dialogue systems is that they need to collect users' information, thus creating the risk of such private data being misused or leaked \cite{krishnamurthy2011privacy, corrigan2014does}. The two configurations (i.e. inter-character and intra-character) we propose in this paper have been implemented in light of this. Being able to understand the impact of each of the profile dimensions within a dialogue system can be useful to determine which are the sensitive data necessary to develop a dialogue system and which could be left out in order to preserve the users' privacy \cite{dudy-etal-2021-refocusing}.
Another problem is the possible fully automated use of profile-based models. Such systems, if left to act completely autonomously, may make erroneous assumptions, even in imitating a given user, thus returning possibly misleading answers. 

\bibliography{custom}

\appendix

\label{sec:appendix}

\newpage

\section{Annotation of PRODIGy Characters Algoritms}

\subsection{Annotation with Personality Information}\label{annotation-mbti-appendix}

Algorithm \ref{annotation-mbti} outlines the annotation process to assign MBTI personality types to the Cornell Movie Dialogue Corpus (CMD). We selected only CMD characters appearing in at least 50 dialogues. For each character we used the query \texttt{movie\_title+year} to extract from the Personality Database (PDB) the related movie metadata, containing the list of the movie characters' names and IDs. If the character was present in the movie metadata, we used a query \texttt{PDB\_characterID} to extract the MBTI type and votes. If the MBTI type has at least 5 votes, the character was annotated. If the character was not found in the movie metadata, a manual check within PDB for character metadata is performed. In case the mismatch could be manually resolved, we replicated the above procedure to annotate the character.

\begin{algorithm}[h]
\small
\caption{MBTI Annotation}
\For{character \textbf{in} CMD characters}{
 \If{nr\_dialogues $\geq$ 50}{
 PDB\_query (movie\_title + year) $\rightarrow$ movie\_metadata\\
 \If{movie\_metadata found}{
 \If{character in movie\_metadata}{
 PDB\_query (PDB\_character\_id) $\rightarrow$ character\_metadata\\
 \If{character\_metadata found}{
 extract MBTI type and n\_votes\\
 \If{n\_votes $\geq$ 5}{
 annotate character\\
 }
 
 }
 \Else{
 manual\_check in PDB $\rightarrow$ character\_metadata\\
 \If{character\_metadata found}{
 extract MBTI type and n\_votes\\
 \If{n\_votes $\geq$ 5}{
 annotate character\\
 }
 }
 }
 }
 }
 }
}

\label{annotation-mbti}
\end{algorithm}

\subsection{Annotation with Biographical Information}\label{annotation-biography-appendix} 

Algorithm \ref{biography-addition-algorithm} describes the process for scraping, revising, and enriching biographies of annotated characters. For each character annotated with MBTI, a biography was scraped from external sources. If a biography was successfully retrieved, an extractive summarisation algorithm based on Kullback-Leibler divergence \cite{haghighi-vanderwende-2009-exploring}  ($KL_{based}$) was applied to extract the most relevant biography sentences and human revision was applied to the sentences.
If no biography was found during the scraping process, the human annotator created a new biography from scratch. Next, an LLM (i.e. ChatGPT) was given the post-edited biography sentences and asked to generate two sets of paraphrased sentences ($sents_{par}$ 1 and $sents_{par}$ 2). Finally, human revision was again applied to the generated sentence sets ($sents_{par}$ 1 and $sents_{par}$ 2), producing the final enriched and revised version of the character's biography.

\begin{algorithm}[h]
\small
\caption{Biographies Scraping, Revision and Enrichment}
\For{character \textbf{in} annotated\_characters}{
 scrape bio from sources\\
 \If{bio exists}{
 \Indp
 $KL_{based}$(bio) $\rightarrow$ bio\_sents\\
 human\_revision(bio\_sents) $\rightarrow$ bio\_sents$_{revised}$\\
 }
 \Else{
 bio\_sents written from scratch
 }
 LLM(bio\_sents$_{revised}$) $\rightarrow$ (sents$_{par}$ 1, sents$_{par}$ 2)\\
 human\_revision(sents$_{par}$ 1, sents$_{par}$ 2) $\rightarrow$ (sents$_{par}$ 1, sents$_{par}$ 2)$_{revised}$\\
}
\label{biography-addition-algorithm}
\end{algorithm}

\section{DialoGPT Fine-tuning Details}\label{dialogpt-training-setup-appendix}

In this section we report the details of the fine-tuning of each model employed during both inter-character and intra-character experiments and the input syntax.
\subsection{Fine-tuning Setup}
To investigate the impact of individual profile dimensions, we opted to employ DialoGPT medium for all fine-tuning experiments. To maintain consistency across our trials, we kept the hyperparameters constant throughout the fine-tuning process, and we considered the type of profile information as the only variable.
In particular, we fine-tuned all our models for 5 epochs with a learning rate of \begin{math}1e-6\end{math} and a batch size of 2. The fine-tuning was performed on a single Tesla V100 GPU.

\subsection{Input Syntax} 
When fine-tuning DialoGPT, we concatenated the characters' profile information to the corresponding turns of the dialogues. The input syntax employed in the experiments conducted with DialoGPT is delineated as follows (we use the example given in Figure \ref{fig:persona_pic} as a reference): \\

\noindent\fbox{ \parbox{2.9in}{ \texttt{<|id|>} u9999 \texttt{<|mbti|>} extrovert, sensor, feeler, perceiver \texttt{<|gender|>} female \texttt{<|bio|>}I am an actress, a star. I live in an old mansion, built for glamorous stars of 1920s Hollywood, just off of Sunset Boulevard. (...) 
\texttt{<|start\_dialogue|>} What's the matter, Norma?\texttt{<|endoftext|>} u9999: Nothing. I just didn't realize what it would be like to come back to the old studio. I had no idea how I'd missed it.\texttt{<|endoftext|>} We've missed you too, dear.\texttt{<|endoftext|>} (...)
u9999: \textit{turn\_to\_be\_predicted}
 } }

 \
\\
\noindent \texttt{<|id|>}, \texttt{<|mbti|>}, \texttt{<|gender|>}, \texttt{<|bio|>} and \texttt{<|start\_dialogue|>} are special tokens added to the model vocabulary, and they are used to segment the input sequence. During fine-tuning, each part of the profile input and its corresponding token are added or removed depending on the configuration under inspection. 

\section{GODEL Prompt Syntax}\label{GODEL-input-syntax-appendix}
During the experiments with GODEL, we prompted the model with an instruction and a context including the profile information and the dialogue context, respectively. We tasked GODEL to predict the last turn in the dialogue.
Following, we provide an example of the input syntax. \\

\noindent\fbox{ \parbox{2.9in}{ Instruction: given a dialog context, you need to respond as a person having the following mbti, gender and bio: "extrovert, sensor, feeler, perceiver", "female", "I am an actress, a star. I live in an old mansion, built for glamorous stars of 1920s Hollywood, just off of Sunset Boulevard. (...)" \texttt{[CONTEXT]} What's the matter, Norma? \texttt{EOS} Nothing. I just didn't realize what it would be like to come back to the old studio. I had no idea how I'd missed it. \texttt{EOS} We've missed you too, dear. \texttt{EOS} (...) \texttt{EOS} \textit{turn\_to\_be\_predicted}
}
}\\

\section{Conditional Perplexity Formulation}\label{cppl-appendix}

Given $T_{n} = \{t_{n_{1}}, t_{n_{2}}, ..., t_{n_{k}} \}$ the $n$th turn with $k$ tokens of a dialogue with history $H=\{T_{1}, T_{2}, ..., T_{n-1}\}$ ($T_{n}$ is the response to $T_{n-1}$), the CPPL of $T_{n}$ is defined as follows: 
\begin{equation}\label{CPPL-eq}
     \textit{CPPL} = \frac{1}{P(T_{n}|H)^{\frac{1}{k}}}
\end{equation} 
    where $P(T_{n}|H)$ is the conditional probability of $T_{n}$ given the history $H$ and $k=|T_{n}|$.

\section{Biography-based Models experiment}\label{bio-experiment}
In order to understand what is the best strategy to input biographies to inter-character models, we conducted a preliminary experiment. In particular,
we tested three strategies to add variability to the biographies during fine-tuning:
(i) \textit{Bio}, trained using the original top-5 biography sentences, (ii) \textit{Bio}$_{rand}$, by randomly selecting, for each dialogue, 5 biography sentences from the corresponding full set of biography sentences of the character, (iii) \textit{Bio}$_{par}$, by randomly selecting 5 sentences for each dialogue from the original biography or from the paraphrases.%

Table \ref{tab:bio-results} shows the effect of randomly choosing 5 sentences out of the full set of biography sentences for each training example (Bio vs. Bio$_{rand}$): randomisation leads to an improvement in terms of \begin{math}CPPL\end{math}. Fine-tuning the models by mixing original and paraphrased biographies, thus increasing lexical variability, improves the performance even further in terms of both \begin{math}CPPL\end{math} (98.27 for Bio$_{par}$ vs. 117.26 for Bio) 
and Acc@\begin{math}N\end{math} (e.g. for Acc@10, 0.661 for Bio$_{par}$ vs. 0.647 for Bio). 
Thus, in the %
inter-character experiments with DialoGPT, we will always use Bio$_{par}$ as the reference configuration.

\begin{table}[h!]
\small
\centering

\begin{tabular}{@{}lrrrrrr@{}}
\toprule
 \textbf{Config.} & \textbf{CPPL} %
 & \textbf{Acc@10} & \textbf{Acc@1} \\ \midrule

Bio & 117.26 %
& 0.647 & 0.294 \\
Bio$_{rand}$ & 106.24 %
& 0.653 & 0.302 \\ 
Bio$_{par}$ & \textbf{98.27} %
& \textbf{0.661} & \textbf{0.307} \\

 \bottomrule
\end{tabular}
\caption{DialoGPT results of the addition of variability to biography sentences on PRODIGy test set (Inter-Character)}
\label{tab:bio-results}
\end{table}

\section{Inter-Character Coherence and Groundedness Analysis}\label{coherence-analysis}

In addition to investigating how different profile dimensions affect \begin{math}CPPL\end{math} and Acc@\begin{math}N\end{math}, we explored their influence on response coherence (i.e. how well the response fits into the conversation) and groundedness (i.e. how relevant the response is based on profile and dialogue information). Results are consistent with  Using \textsc{UniEval} by \citet{zhong-etal-2022-towards}, we assessed coherence and groundedness of responses from models trained on individual profile dimensions, alongside gold responses. Our analysis (Table \ref{tab:profile-coherence}) shows that: (i) all profile-based models have better metrics than plain dialogue; (ii) gold responses are the most coherent and relevant, highlighting room for improvement for our models. Among our models, the Gender model yields the most coherent responses (0.526), while the Bio$_{par}$ model generates the most grounded responses (0.057).

\begin{table}[ht!]
\small
\centering
\begin{tabular}{l|r|r}
\toprule
                              & \textbf{Coherence} & \textbf{Groundedness} \\ \midrule
Gold                          & 0.581     & 0.066        \\\midrule
\Hermaphrodite & \textbf{0.526}     & 0.037        \\
\texttt{MBTI}                          & 0.520     & 0.033        \\
Bio$_{par}$                           & 0.507     & \textbf{0.057}        \\
PD                            & 0.462     & 0.026       \\\bottomrule
\end{tabular}
\caption{Evaluation of coherence and groundedness scores for model-generated responses compared to gold standard responses. The scoring range is [0, 1].}
\label{tab:profile-coherence}
\end{table}

\section{Analysis of Human Evaluation Rankings}\label{human-evaluation-appendix}
Table \ref{tab:human-eval-rankings} presents the evaluators' average rankings. The scores are inverted for readability purposes: higher scores indicate better performances. The significant gap between the scores of gold and the generated responses indicates that there is wide room for improvement for our models. Among the models, Plain Dialogue receives the highest ratings, closely followed by the other models. In shorter contexts, profile-based models, i.e., Bio$_{par}$, MBTI, Gender, yield higher scores than in longer context: this suggests that profile information is beneficial when dialogue context does not provide sufficient information about the speaker. %
Furthermore, when the profile information is explicitly provided to evaluators, the gap between scores in shorter and longer dialogues diminishes. This suggests a positive impact of profile information on evaluators' judgements, who perceive responses generated by profile-based models as more appropriate.

\begin{table}[ht!]
\small
\centering
\begin{tabularx}{\linewidth}{l *{6}{>{\centering\arraybackslash}X}}
\toprule
 & \multicolumn{2}{r}{\textbf{All turns}} & \multicolumn{2}{r}{\textbf{$\leq $ 6 turns}} & \multicolumn{2}{r}{\textbf{$>$ 6 turns}} \\
\cmidrule(lr){2-3} \cmidrule(lr){4-5} \cmidrule(lr){6-7}
\textbf{Response}& \textbf{No \includegraphics[height=1em]{profile_pic.png}} & \textbf{With \includegraphics[height=1em]{profile_pic.png}} & \textbf{No \includegraphics[height=1em]{profile_pic.png}} & \textbf{With \includegraphics[height=1em]{profile_pic.png}} & \textbf{No \includegraphics[height=1em]{profile_pic.png}} & \textbf{With \includegraphics[height=1em]{profile_pic.png}} \\
\midrule
Gold & \textbf{4.04} & \textbf{3.97} & \textbf{3.90} & \textbf{3.85} & \textbf{4.32} & \textbf{4.18} \\
PD & 2.90 & 2.86 & 2.89 & 2.89 & 2.92 & 2.80 \\
Bio$_{par}$ & 2.66 & 2.71 & 2.70 & 2.77 & 2.58 & 2.60 \\
MBTI & 2.67 & 2.75 & 2.77 & 2.77 & 2.49 & 2.70 \\
\Hermaphrodite & 2.73 & 2.71 & 2.75 & 2.71 & 2.69 & 2.72 \\
\bottomrule
\end{tabularx}
\caption{Human Evaluation Average Rankings across different dialogue lengths: higher scores indicate better performance. PD and \Hermaphrodite represent Plain Dialogue and Gender, respectively. No/With \includegraphics[height=1em]{profile_pic.png} indicates profile information disclosure to evaluators.}
\label{tab:human-eval-rankings}
\end{table}

\section{Intra-Character Generations examples}

In Table \ref{tab:example-generations-intra}, we provide a few examples of model generations derived from the Intra-Character configuration. Similar to the Inter-Character setup, the Plain Dialogue model produces a generic response that seamlessly fits the dialogue context. However, it is noteworthy that the output from the Bio model is particularly aligned with the Gold response concept.

\begin{table}[h]
\centering
\small
\begin{tabular}{p{0.45\textwidth}}
\toprule
\textbf{\texttt{Input}}  \\
\begin{minipage}[t]{0.45\textwidth}
\textbf{Biography}: (1) I live in what is left of Philadelphia the year of 2035. (2) I am single, but I am developing feelings for my psychiatrist, Dr. Railly. (3) I am a total nutcase, completely deranged, delusional, paranoid. (4) I am visiting 1990, after being tossed in jail, I have been sent back in time to learn about a virus before it started mutating. (5) When I told others I am from the future, they thought I was out of my mind and threw me in a insane asylum. 
\textbf{\texttt{MBTI}}: Extrovert, Intuititive, Feeler, Perceiver \\

\textbf{\texttt{Gender}}: Male \\
\vspace{0.2cm}
\hrule
\vspace{0.2cm}
\textbf{\texttt{Dialogue}:}\\
  \textbf{A:} What have you done?  Did you...kill someone? \\
  \textbf{B:}  No!  I...don't think so.  I stole a car and they chased me.  I hit a tree. \\
  \textbf{A:} See -- you can drive after all! \\
\vspace{0.1cm}
\end{minipage}
\vspace{0.2cm}
\hrule
\vspace{0.2cm}
 \textbf{\texttt{Responses}}
\begin{minipage}[t]{0.45\textwidth}
\begin{itemize}[leftmargin=*]
\itemsep0em
  \item \textbf{[\texttt{Gold}]} \textbf{B:} Yeah, sort of, I guess.  I...I'm sorry I locked you up.  I thought...I thought... I think maybe I am crazy!
  \item \textbf{[\texttt{Plain Dialogue}]} \textbf{B:} I don't want to!
  \item \textbf{[\texttt{Char ID}]} \textbf{B:} Yeah!  Yeah!!  So what if we killed the cops!  Now, I need to meet you in there tomorrow morning!
  \item \textbf{[\texttt{Bio}]} \textbf{B:} Oh, God. I must be totally crazy.
  \item \textbf{[\texttt{MBTI}]} \textbf{B:} Yeah, but, what are you gonna do?
  \item \textbf{[\texttt{Gender}]} \textbf{B:} ...and I thought the cops were gonna catch me.  They're not gonna catch me.
\end{itemize}
\end{minipage} \\
\bottomrule
\end{tabular}
\caption{Example of Intra-Character models' generations. }
\label{tab:example-generations-intra}
\end{table}

\end{document}